\documentclass[runningheads]{llncs}

\usepackage[mobile]{eccv}

\usepackage{eccvabbrv}

\usepackage{graphicx}
\usepackage{booktabs}
\usepackage{kotex}
\usepackage{enumitem}
\usepackage{mathtools}

\usepackage[accsupp]{axessibility}  % Improves PDF readability for those with disabilities.

\usepackage{comment}
\usepackage{wrapfig}

\usepackage{hyperref}

\usepackage{orcidlink}

\usepackage{amsmath,amssymb}

\begin{document}

\title{Harmonizing Visual and Textual Embeddings \\ for Zero-Shot Text-to-Image Customization
}

\titlerunning{Harmonizing Visual and Textual Embeddings}

\author{Yeji Song\inst{1} \and
Jimyeong Kim\inst{1}$^{\ast}$ \and
Wonhark Park\inst{1}$^{\ast}$ \and
Wonsik Shin\inst{1} \and \\
Wonjong Rhee\inst{1} \and
Nojun Kwak\inst{1}$^{\dagger}$}

\authorrunning{Yeji Song et al.}

\institute{Seoul National University
\email{\{ldynx,wlaud1001,pwh0515,wonsikshin,wrhee,nojunk\}@snu.ac.kr}}

\maketitle

\let\thefootnote\relax\footnotetext{
$\ast$ Equal contribution \quad $\dagger$ Corresponding author
}

\begin{abstract}
In a surge of text-to-image (T2I) models and their customization methods that generate new images of a user-provided subject, current works focus on alleviating the costs incurred by a lengthy per-subject optimization. These zero-shot customization methods encode the image of a specified subject into a visual embedding which is then utilized alongside the textual embedding for diffusion guidance. The visual embedding incorporates intrinsic information about the subject, while the textual embedding provides a new, transient context. However, the existing methods often 1) are significantly affected by the input images, \eg, generating images with the same pose, and 2) exhibit deterioration in the subject's identity. We first pin down the problem and show that redundant pose information in the visual embedding interferes with the textual embedding containing the desired pose information. To address this issue, we propose \textit{orthogonal visual embedding} which effectively harmonizes with the given textual embedding. We also adopt the visual-only embedding and inject the subject's clear features utilizing a \textit{self-attention swap}. Our results demonstrate the effectiveness and robustness of our method, which offers highly flexible zero-shot generation while effectively maintaining the subject's identity. The code will be publicly released.
\end{abstract}

\section{Introduction}
\label{sec:intro}

Recent advancements in text-to-image (T2I) generation, especially diffusion models~\cite{rombach2022ldm, balaji2023ediffi, saharia2022imagen,ramesh2022dalle2,nichol2022glide} have opened up a new era of image creation. 
Subject-driven generation~\cite{ruiz2023dreambooth,gal2022textual_inversion,tewel2023perfusion,qiu2023controlling,voynov2023p} aims to generate novel images featuring a specific subject provided by the user. 
The common approach represents the subject as a new pseudo-word ($\text{S}^*$) in the textual embedding space of the text encoder. 
They optimize a pseudo-word by updating the textual embedding of the pseudo-word~\cite{gal2022image} or the diffusion model's parameters~\cite{ruiz2023dreambooth,tewel2023key}. 
However, both approaches require per-subject optimization, posing challenges for real-time applications and multi-subject customization.
In response to this challenge, zero-shot customization methods~\cite{li2024blip,wei2023elite,chen2023photoverse,shi2023instantbooth,ma2023subjectdiffusionopen,jia2023tamingencoder,yuan2023customnet,xiao2023fastcomposer} have been proposed. 

They adopt the mappers (\eg, MLP network~\cite{wei2023elite,yuan2023customnet,xiao2023fastcomposer}, adapter~\cite{chen2023photoverse,shi2023instantbooth,ma2023subjectdiffusionopen} or multi-modal encoder~\cite{li2024blip}) to transform a subject's image into the visual embedding, which is subsequently utilized along with the textual embedding. 
The visual embedding provides representative information about the subject's identity from the input image, eliminating the need for additional training processes. 
To improve the user's convenience, most zero-shot customization methods~\cite{li2024blip,wei2023elite,chen2023photoverse,ma2023subjectdiffusionopen,jia2023tamingencoder,yuan2023customnet,xiao2023fastcomposer} encode a single image as the input. In line with this practice, we focus on single-image-based zero-shot customization in this paper.

\begin{figure}[t!]
    \centering
    \includegraphics[width=\textwidth]{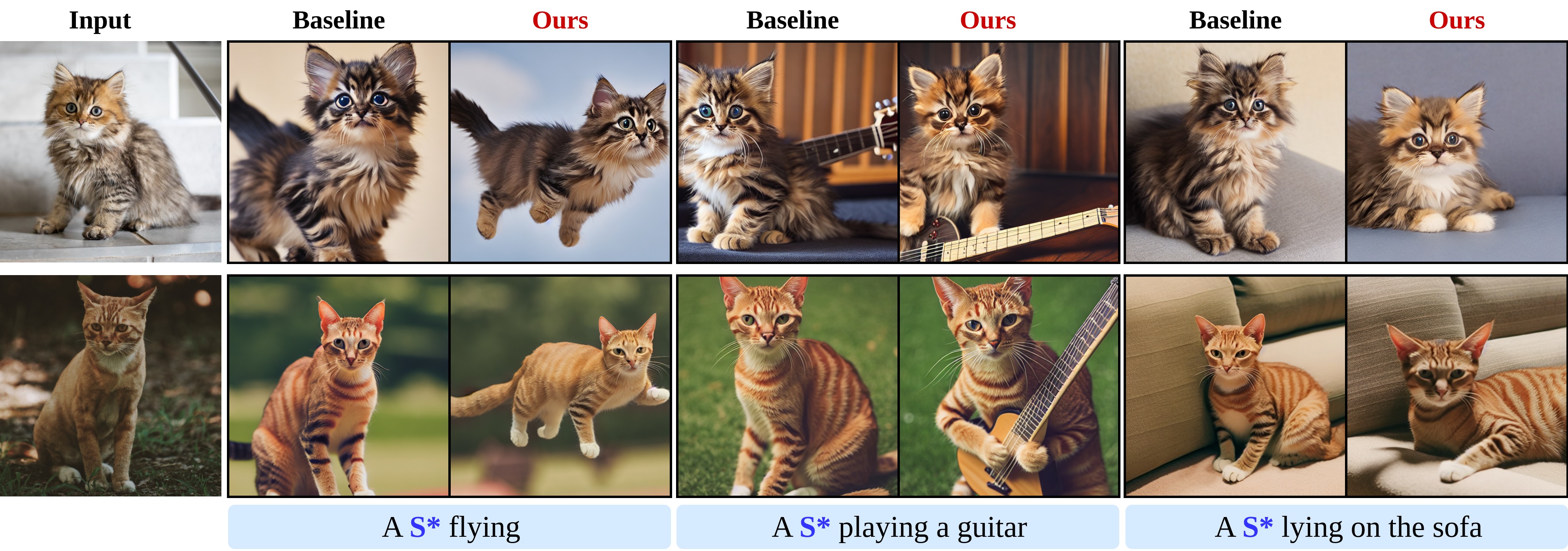}
    \caption{Our method deals with the challenges of pose variation in zero-shot customization methods \ie, (i) strong adherence to the pose in the input image and (ii) loss in the subject’s identity. Our method paves the way for a more diverse, efficient, and lively subject-driven generation.
    }
    \vspace{-1em}
    \label{fig:intro}
\end{figure}

When dealing with a single image, the existing methods focus on separating the subject's identity from other irrelevant details within an image, as the input image typically encompasses more than just the subject's identity. To address this challenge, they employ the subject's segmentation mask~\cite{li2024blip,wei2023elite,ma2023subjectdiffusionopen,jia2023tamingencoder,xiao2023fastcomposer} or the visual embedding from the deepest layers of an encoder~\cite{wei2023elite}. % to separate its identity from the rest of the image. 
% \jimyeong{While these techniques effectively separate the subject from the background or other objects within the image, %the pose information still remains.
% \yeji{it cannot disentangle the subject's pose from its identity as these aspects are tightly intertwined within the same pixels. }
% Consequently, this information may inevitably seep into the visual embedding, inducing conflict between visual and textual embeddings when the text prompt is given to generate a subject with a pose differing from that of the input image. 
% We have identified two significant problems that stem from this conflict }\yeji{namely the (i) \textit{pose bias} and (ii) \textit{identity loss}:}}
% Orig:
However, while those techniques effectively separate the subject from the background or other objects within the image, disentangling the subject's pose from its identity is almost impossible as these aspects are tightly intertwined within the same pixels. We term this inevitable phenomenon as the \textit{pose-identity entanglement} within visual embedding, which becomes apparent when attempting to change the pose of the subject while preserving the subject's identity. Regarding this, we have identified two significant problems when modifying the subject's pose with a text prompt, namely the (i) \textit{pose bias} and (ii) \textit{identity loss}: 
\begin{enumerate}[label=(\roman*)]
\item The generated images tend to either maintain the original pose of the subject or exhibit some unnaturalness that falls between the pose derived from the text prompt and the one presented in the input image.
\item The subject in the generated images partially loses its identity, appearing with a different color or body shape.
\end{enumerate}
In this paper, we focus on resolving the pose bias and identity loss described above. They are imperative tasks for advancing towards a more diverse and desirable customization, while highly challenging due to the pose-identity entanglement.
We start from unveiling the more direct and addressable cause of the two problems. We found that using only the textual embedding generates the subject in various poses faithfully complying with the text prompt (without \textit{pose bias}) while using only the visual embedding results in the images where the subject retains its whole identity (without \textit{identity loss}). This leads us to assume that a conflict exists between the visual and textual embeddings, from which the two problems mainly originate. We provide the result images in the supplementary materials. Regarding pose-identity entanglement, the input image carries the visual features of \textit{the subject in a specific pose}.
Therefore, when generating a novel image of the subject in a different pose, the input image and text prompt provide two concurrent but conflicting embeddings to the model with different pose information.
As a result, the visual embedding readily interferes with the textual embedding, causing the \textit{pose bias}.
Conversely, textual embedding can also interfere with visual embedding, affecting the subject's identity and resulting in \textit{identity loss}.

To alleviate this conflict, we propose \textbf{contextual embedding orchestration} and \textbf{self-attention swap} that effectively resolve each problem. 
The former involves adjusting the visual embedding to align better with the textual embedding. This adjustment is achieved by orthogonalizing the visual embedding vector to the subspace of the textual embedding vectors, essentially reducing the interference of the visual embedding. The latter adopts another denoising process guided by the visual-only embedding that fundamentally evades the collision between the visual and textual embedding and aggregates clean information about the subject. By swapping the self-attention between the original and the visual-only-guided diffusion models, we could successfully maintain the subject's identity in newly generated images. 
Our method is generic and easily applicable to any zero-shot customization method that utilizes visual and textual embeddings, as it does not require an additional tuning process.
We demonstrate that our method significantly improves the pose variation of the baseline without compromising its performance in pose-irrelevant scenarios, \eg, placement of novel objects or changing the subject's texture. \cref{fig:intro} illustrates that our method effectively resolves the pose bias. We also provide the results regarding identity loss in \cref{fig:qualitative,fig:qualitative_elitr}, and supplementary materials.

Our contributions are summarized as:
\begin{itemize}
    \item[$\bullet$] We unveil the problem of pose bias and identity loss in zero-shot customization and shed light on the discord among the visual and textual embeddings. 
    \item[$\bullet$] Our proposed method effectively resolves the pose bias and identity loss, offering highly diverse and pose-variant subject generation.
    \item[$\bullet$] Our method is readily applicable to any zero-shot customization methods that incorporate visual and textual embeddings.
\end{itemize}

\section{Related Works}

\textbf{Text-to-Image Generation.}
Amidst a proliferation of image synthesis models, diffusion models~\cite{sohldickstein2015deep,ho2020ddpm}  have demonstrated their strength in producing images with remarkable fidelity and comprehensive mode coverage. Their capacity to generate a spectrum of diverse images has facilitated their integration with large pre-trained language models~\cite{radford2021clip}. This synergy has given rise to diffusion-based T2I models~\cite{rombach2022ldm, balaji2023ediffi, saharia2022imagen,ramesh2022dalle2,nichol2022glide}, which can generate high-quality images with strong controllability by the guidance of natural language instructions. Recently, to increase the flexibility using this strong prior, many have combined conditioning embeddings from different modalities, \eg, concatenating text-aligned visual embedding extracted from an image with textual embedding. 
However, simply combining various embeddings may cause conflict in dealing with different information they have. Hence, our primary goal is to devise an appropriate methodology for integrating diverse embeddings from different modalities, considering potential discrepancies in their inherent information. 

\medskip
\noindent \textbf{Subject-driven Generation.} Given a few images of a user-provided subject, subject-driven generation~\cite{ruiz2023dreambooth,gal2022textual_inversion,tewel2023perfusion,qiu2023controlling,voynov2023p} aims to generate images containing the subject in various contexts instructed by text guidance. However, per-subject optimization suffers from computation and memory burden, leading to an introduction of zero-shot customization methods~\cite{li2024blip,wei2023elite,chen2023photoverse,shi2023instantbooth,ma2023subjectdiffusionopen,jia2023tamingencoder,yuan2023customnet}. They involve pre-training the mapper enabling the transformation of the input image into text-aligned visual embeddings without per-subject optimization. The visual embedding is then concatenated with textual embedding and utilized as guidance to the T2I models. However, we found that they often struggle with the interference between the visual and textual embeddings, often failing to properly deliver text prompts. Our proposed method first pins down the interference between the two embeddings and provides a solution for it, enabling more diverse and more flexible image generations.

Meanwhile, to edit the context while maintaining the subject and the background, \textit{consistent} image editing methods~\cite{cao_2023_masactrl,kawar2023imagic} have been proposed. While the general image editing methods~\cite{hertz2022prompt,nichol2022glide,parmar2023pix2pix_zero,yang2022paint, Avrahami_2022blendeddiffusion,couairon2022diffedit,mokady2022nulltext,tumanyan2022plugandplay,zhang2023controlnet,mou2023t2iadapter} change an image style, surrounding objects, a subject's appearance, or a subject per se, consistent methods target to change the subject's pose or viewpoint while preserving the subject's identity. Compared to customization methods that deal with noise-to-image generations, those methods focus on image-to-image translation. They edit the subject without changing the other context, \eg, maintaining the input image's structure, background, and other surrounding objects. 
On the other hand, customization methods aim to synthesize the subject in diverse context according to the provided text prompt. Our method concentrates on customization while incorporating the image editing capability to address issues stemming from zero-shot customization.

\medskip
\noindent \textbf{Compositional Generation.} Due to the limited size of the textual embeddings, large pre-trained T2I diffusion models suffer from fully compositing complex text descriptions~\cite{liu2023compositional}. Precedent research tackles this problem and offers various solutions; \eg, giving an additional segment mask layout for each related prompt as a condition~\cite{kim2023dense}, composing separate diffusion models where each is encoded with a divided prompt~\cite{liu2023compositional}, and using the perpendicular gradient as a negative prompt guidance~\cite{armandpour2023perpneg}. Likewise, we deal with complex prompts consisting of visual and textual embeddings and suggest how to convey information without interference between them.

\section{Preliminaries}

In this work, we employ text-to-image latent diffusion model (LDM)~\cite{rombach2022ldm}. 
The denoising process is implemented in the latent space using the autoencoder structure with the encoder $\mathcal{E}(\cdot)$ and the decoder $\mathcal{D}(\cdot)$. Specifically, an image $x$ is projected to a latent representation $z=\mathcal{E}(x)$, and decoded back to the image space giving $\Tilde{x}=\mathcal{D}(z)=\mathcal{D}(\mathcal{E}(x))$, reconstructing $x$, \ie, $\Tilde{x} \approx x$. Given the pre-trained autoencoder, the latent diffusion model $\epsilon_\theta(x_\tau,\tau); \tau=1 \cdots T$ is trained with the following objective where $\textbf{c}$ represents the contextual embedding of textual/visual condition generally obtained from the pre-trained CLIP encoder~\cite{radford2021clip}:
\begin{equation} \label{eq:ldm_train}
    L_{LDM} = \mathbb{E}_{x \sim p(x), \epsilon \sim \mathcal{N}(0,I),\textbf{c},\tau \sim \text{uniform}(1,T)}[\| \epsilon - \epsilon_\theta(z_\tau(x),\tau,\textbf{c}) \|^2_2]. 
\end{equation}
During inference, $z_T \sim \mathcal{N}(0,I)$ is iteratively denoised to the initial representation $z_0(x) = \mathcal{E}(x)$.

Text prompts act as a condition on LDM through the cross-attention mechanism. The latent spatial feature $f \in \mathbb{R}^{l \times h}$ is projected to produce the \textit{query} $Q= f \cdot W_Q  \in \mathbb{R}^{l \times d}$ while the text prompts are first encoded into the text embedding $\textbf{c} \in \mathbb{R}^{l_c \times h_c}$ and projected to yield the \textit{key}, $K=\textbf{c} \cdot W_K \in \mathbb{R}^{l_c \times d}$, and \textit{value}, $V=\textbf{c} \cdot W_V \in \mathbb{R}^{l_c \times d}$. Then the cross-attention is calculated as follows:
\begin{equation} \label{eq:attention}
    \text{Attention}(Q,K,V) = \text{softmax}(QK^T / \sqrt{d}) \cdot V
\end{equation}
where $l$, $l_c$, $d$ and $h$ are spatial sequence length, context sequence length, dimension of key/query/value, and dimension of spatial feature/context respectively.
Self-attetntion mechanism uses the same \cref{eq:attention} where the key and value are attained from latent spatial features $f$ instead of the contextual embedding $\textbf{c}$.

\section{Methods}

\subsection{Discord among Contextual embeddings} \label{subsec:discord}

\begin{figure}[t!]
    \centering
    \includegraphics[width=\textwidth]{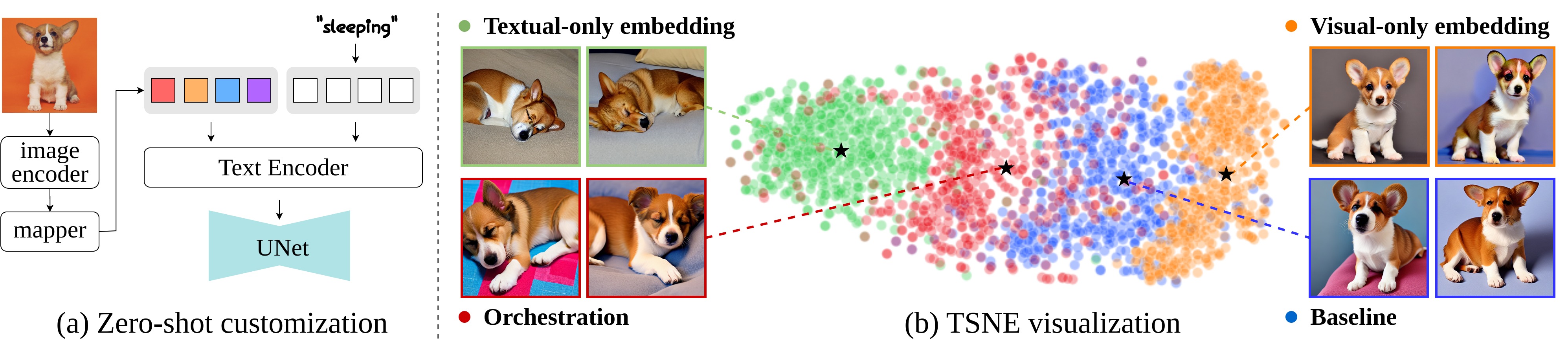}
    \caption{\textbf{Visualization of discord among contextual embeddings.} (a) The contextual embedding of zero-shot customization methods is generally composed of visual and textual embedding, each of which provides the subject's identity and a novel context, respectively. (b) However, pose information in the visual embedding interferes with the textual embedding, resulting in biased images that are closely attached to the input image. Meanwhile, our orchestration resolves the conflict, generating images that are both identity-conservative and faithful to the text prompt.}
    \vspace{-1em}
    \label{fig:discord}
\end{figure}

Single-image-based zero-shot customization methods~\cite{li2024blip,wei2023elite,chen2023photoverse,ma2023subjectdiffusionopen,jia2023tamingencoder,yuan2023customnet,xiao2023fastcomposer} compose the contextual embedding by concatenating two heterogeneous embeddings; visual embedding $\textbf{v} \in \mathbb{R}^{M \times h_c}$ and textual embedding $\textbf{t} \in \mathbb{R}^{N\times h_c}$ ($\textbf{c} = [\textbf{v};\textbf{t}] = [v_1 \cdots v_M; t_1 \cdots t_N]$).
Image features of a given subject's image are transformed into the visual embedding using the pre-trained mapper. The visual embedding is then combined with the textual embedding, engaging in the spatial features $z$ via cross-attention layers as in \cref{eq:attention} or additional adapters. 
We depict the overall zero-shot pipeline in ~\cref{fig:discord}(a). 

Meanwhile, when modifying the subject's pose with a text prompt, the visual embedding, which includes pose information from the input image, conflicts with the textual embedding, rendering the images confined to the specific pose or partially losing the subject's identity. identity of the subject's concept. To verify the interference between the visual and textual embedding, we conduct an experiment utilizing three types of contextual embedding respectively: 1) contextual embedding composed solely of the visual embedding ($\textbf{c}_{\textbf{v},\varnothing} = [v_1 \cdots v_M; \varnothing]$), 2) solely of the textual embedding ($\textbf{c}_{\varnothing,\textbf{t}} = [\varnothing; t_1 \cdots t_N]$), and 3) the original combined contextual embedding ($\textbf{c} = [v_1 \cdots v_M; t_1 \cdots t_N]$). We generate 500 images conditioned on the respective contextual embedding and obtain their representations from the highest layer of VGG-16~\cite{simonyan2014very} as visualized in \cref{fig:discord}(b). Ideally, the generated images conditioned on the original contextual embedding should include the subject's identity and, at the same time, faithfully follow the text prompts. However, their image features exhibit a strong bias toward outputs from the visual-only embedding.
The pre-existing pose information in visual embedding interferes with the textual embedding, restricting the generated images to be closely tied to the visual embedding, \ie, maintaining the pose of the subject in the input images instead of constructing a novel pose as the textual embedding instructs. The result above clarifies that the visual embedding interferes with the textual embedding, confining the generated images away from text guidance. Conversely, the textual embedding also affects the visual embedding, harming the subject's identity. We include the analysis regarding the identity loss in supplementary materials. These results underline the necessity of adjusting the visual embedding to resolve this discord.

\subsection{Contextual embedding Orchestration}
\label{subsec:orch}
To resolve the interference between the visual embedding $\textbf{v}$ and the textual embedding $\textbf{t}$ as presented in \cref{subsec:discord}, we propose \textbf{orthogonal visual embedding} $\textbf{v}^\bot$ as follows:
\begin{equation} \label{eq:orthogonal}
v^\bot = v- v^{||} = v - \! \sum_{j=1,  j \neq \text{sbj}}^{N} \langle \bar{t}_j, v \rangle \bar{t}_j , \quad \bar{t}_j = \text{normal}\left(t_j - \! \sum_{i=1, i\neq \text{sbj}}^{j-1} \langle \bar{t}_i, t_j \rangle \bar{t}_i \right)
\end{equation}
where $t_{\text{sbj}}$ represents the textual embedding corresponding to the pseudo-word (\eg, `S$^*$' in Fig. \ref{fig:intro}) and $\text{normal}(\cdot)$ means $l_2$ normalization\footnote{We incorporate all text tokens except for $\text{S}^*$ and its class name since they are closely related to pose \eg, when we generate an image of ``a $\text{S}^*$ eating a hamburger'', the token ``hamburger'' also affects the eating pose and should be considered, too.}.
$\{\bar{t}_j \}_{j\neq \text{sbj}}$ are the basis vectors of the textual subspace, obtained by the Gram-Schmidt orthogonalization process.
\cref{eq:orthogonal} can be viewed as breaking down a vector $v$ into two components, $v^\bot$ and $v^{||}$, of which $v^{||}$ resides in the textual subspace which is spanned by $\{\bar{t}_j \}_{j\neq \text{sbj}}$.

We argue that $v^{||}$ causes the interference with $\textbf{t}$ when presented concurrently. 
The other component, $v^\bot$, is perpendicular to the textual subspace, embodying the essential information about the subject's identity alleviating the interference with $\textbf{t}$. 
Using $v^\bot$ instead of $v$, we could establish the new axes in the visual embedding that interplays more effectively with $\textbf{t}$, helping resolve the interference.
The new contextual embedding $\textbf{c}^\bot = [v^\bot_1 \cdots v^\bot_M; t_1 \cdots t_N]$ is then incorporated with latent spatial features $f$ via cross-attention. Note that we also exclude the textual embedding of articles or the subject's class name from \cref{eq:orthogonal} when such tokens are used.
~\cref{fig:method}(a) illustrates the contextual embedding orchestration where the textual embedding $\textbf{t}$ effectively provides guidance with alleviated interference from the visual embedding $v^\bot$.

\subsection{Self-attention Swap}

We found that the conflict between visual and textual embeddings results in the loss of the subject's identity, when the textual embedding affects the visual embeddings. This phenomenon is still visible even after orchestration, due to the inevitable pose-identity entanglement; changing the visual embedding inevitably results in a change in identity. To alleviate the loss in the subject's identity, we propose a self-attention swap to retain essential information about the subject. 
Our intuition is based on the observation that when the visual-only embedding is used as the contextual embedding ($\textbf{c}_{\textbf{v},\varnothing} = [v_1 \cdots v_M; \varnothing]$), the output images are clean, effectively preserving the subject's identity, as it is free of the interference of the textual embedding.

\begin{figure}[t!]
    \centering
    \includegraphics[width=\textwidth]{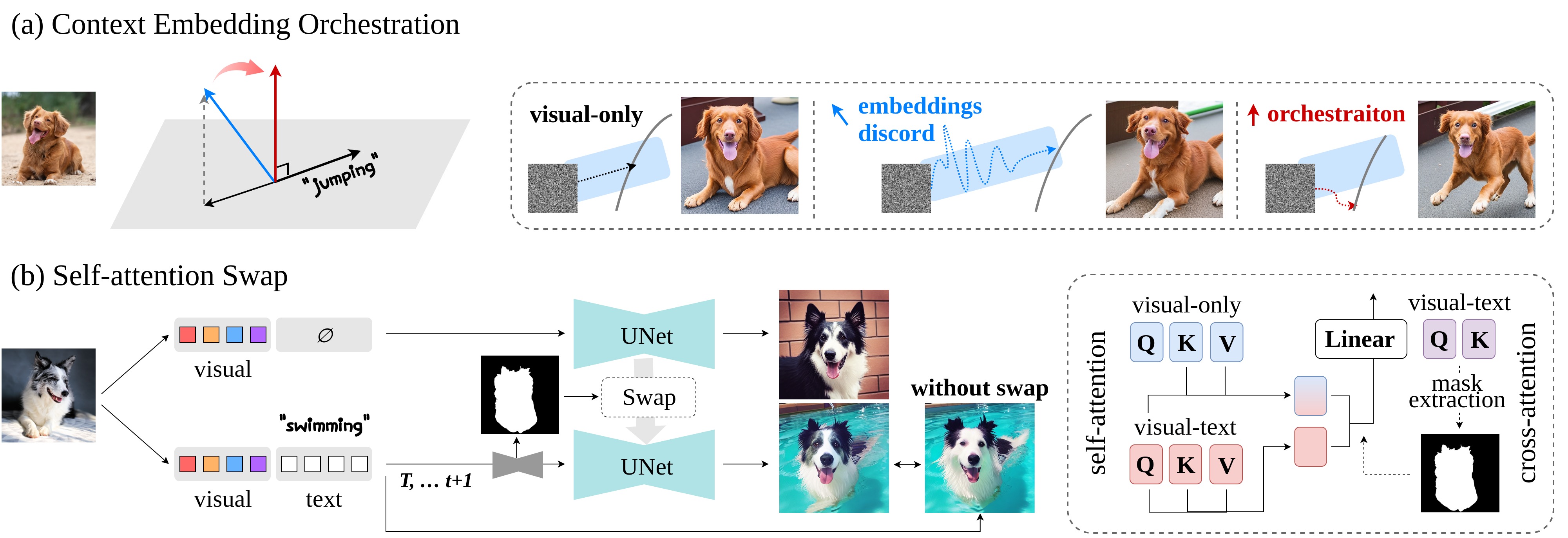}
    \caption{\textbf{The overall pipeline of our method}. (a) To alleviate the pose-bias due to the visual embedding, we conduct \textit{Orchestration} of the contextual embedding \ie, adjust the visual embedding to be orthogonal to the textual embedding. Orchestration accurately guides the denoising process toward a direction that follows the pose directed by the text prompt. (b) \textit{Self-attention Swap} obtains self-attention key and value from another denoising process guided by visual-only embedding, which offers the subject's clean identity. After swapping, our method retains the subject's identity while changing its pose faithfully following the text prompt.}
    \vspace{-1em}
    \label{fig:method}
\end{figure}

To effectively attain the subject's identity using the visual-only embedding while simultaneously maintaining a novel pose of the subject, we delve into the image editing methods. The image editing methods generally employ two denoising processes, one for a source image reconstruction and another for generating a new, edited image. 
The properties of the source image obtained during its reconstruction are transferred to the process of generating the edited image by swapping self-attention~\cite{cao_2023_masactrl,tumanyan2022plugandplay} or cross-attention~\cite{hertz2022prompt,parmar2023pix2pix_zero,couairon2022diffedit}.

Inspired by them, we adopt the second denoising process $\{z'_\tau\}_{\tau=1 \cdots T}$ where the visual-only embedding is provided as the contextual embedding ($\textbf{c}_{\textbf{v},\varnothing} = [v_1 \cdots v_M; \varnothing]$). The original denoising process is denoted as $\{z_\tau\}_{\tau=1 \cdots T}$ that utilizes orthogonal visual embedding in \cref{eq:orthogonal} along with the textual embedding ($\textbf{c}^\bot = [v^\bot_1 \cdots v^\bot_M; t_1 \cdots t_N]$). Then, we modify the self-attention layers of $\{z_\tau\}_{\tau=1 \cdots T}$, to swap key and value with those from $\{z'_\tau\}_{\tau=1 \cdots T}$. Our proposed self-attention swap can be formulated as follows:
\begin{equation} \label{eq:swap}
\text{AttnSwap}(z, z') = M'_S V', \hspace{0.4em} \text{where} \hspace{0.4em} M'_S = \text{softmax}(QK'^T/\sqrt{d}).
\end{equation}
Here, $Q \in \mathbb{R}^{l \times d}$ is the query from $\{z_\tau\}_{\tau=1 \cdots T}$ and $K', V' \in \mathbb{R}^{l \times d}$ are the key and the value from $\{z'_\tau\}_{\tau=1 \cdots T}$ where $l$ and $d$ are spatial sequence length and features dimension, respectively. $M'_S \in \mathbb{R}^{l \times l}$ evaluates how closely connected latent spatial features from $z_t$ and $z'_t$ are, assigning higher scores to latent pixels when they have a strong relationship. 
Using \cref{eq:swap}, we inject $V'$ in each latent pixel from $\{z'_\tau\}_{\tau=1 \cdots T}$ based on $M'_S$. 
In other words, while generating a novel pose of the subject, we incorporate \textit{values} from the clear identity of the subject into the location where the corresponding latent pixels exist.

As we aim to preserve the subject's identity while allowing flexibility in the remaining aspects, we restrict self-attention swap within latent pixels assigned to the subject. Inspired by \cite{hertz2022prompt,cao_2023_masactrl}, we utilize the cross-attention maps $M_C$ from the original denoising process to create a mask distinguishing the subject from the background. $M_C \in \mathbb{R}^{l \times l_c}$ indicates how much information of text tokens is associated with each latent pixel where $l_c$ is the context sequence length. We could obtain a binary mask $m \in \{0,1\}^l$ for the subject token with fixed thresholds. We ultimately adopt self-attention swap as follows:
\begin{equation} \label{eq:swap2}
\text{MaskedAttnSwap}(z, z') = \text{AttnSwap}(z, z') * m + M_S V * (1 - m),
\end{equation}
where $M_S = \text{softmax}(QK^T / \sqrt{d})$ and $K, V$ are key and value features from the original process $\{z_\tau\}_{\tau=1 \cdots T}$. \cref{fig:method}(b) shows the overall process of self-attention swap. We also provide more details in the supplementary materials.

\section{Experiments}

\noindent \paragraph{Datasets.} 
Since prevailing benchmark datasets~\cite{ruiz2023dreambooth, kumari2023multi} primarily utilize generating prompts related to changing backgrounds or introducing new objects, they often fall short in effectively evaluating the crucial aspect of modifying subject poses.
To address this limitation, we have constructed a new dataset, \textbf{Deformable Subject Set} (DS set), to effectively assess the model's capability to modify a subject's pose.
The DS set comprises 19 live animals from the DreamBooth~\cite{ruiz2023dreambooth} and CustomDiffusion~\cite{kumari2023multi}, along with 11 prompts specifically designed to focus on the deformation of the subjects' poses.
Furthermore, we also utilized the \textbf{DreamBooth dataset} (DB set)~\cite{ruiz2023dreambooth} to evaluate the model's capacity in typical scenarios.

\noindent \paragraph{Metrics.} 
Following DreamBooth~\cite{ruiz2023dreambooth}, we measured the subject fidelity using CLIP-I and DINO-I, and measured text alignment using CLIP-T. 
For the DS set, we additionally measured the masked scores for CLIP-I and DINO-I, \textit{incorporating the subject's segmentation mask} for both the input images and generated images, as performed in \cite{avrahami2023break}.
We integrated segmentation masks into our image alignment measures to mitigate the impact of newly introduced objects in generated images, allowing for a more focused comparison of the subject's identity. 
This approach is particularly valuable when prompts necessitate the addition of new objects alongside the subject in the image.

\noindent \paragraph{Implementation Details.}
In our main experiment, we employed open-sourced zero-shot customization models, BLIP-Diffusion~\cite{li2024blip} and ELITE~\cite{wei2023elite} as the base models. %for zero-shot customization.
We utilized the PNDM scheduler~\cite{liu2022pseudo} for Blip-Diffusion and the LMS scheduler~\cite{karras2022elucidating} for ELITE, each with 100 denoising steps, according to their default settings.
Self-attention swap was applied after the 20th denoising step.

\begin{figure}[t!]
    \centering
    \includegraphics[width=.9\textwidth]{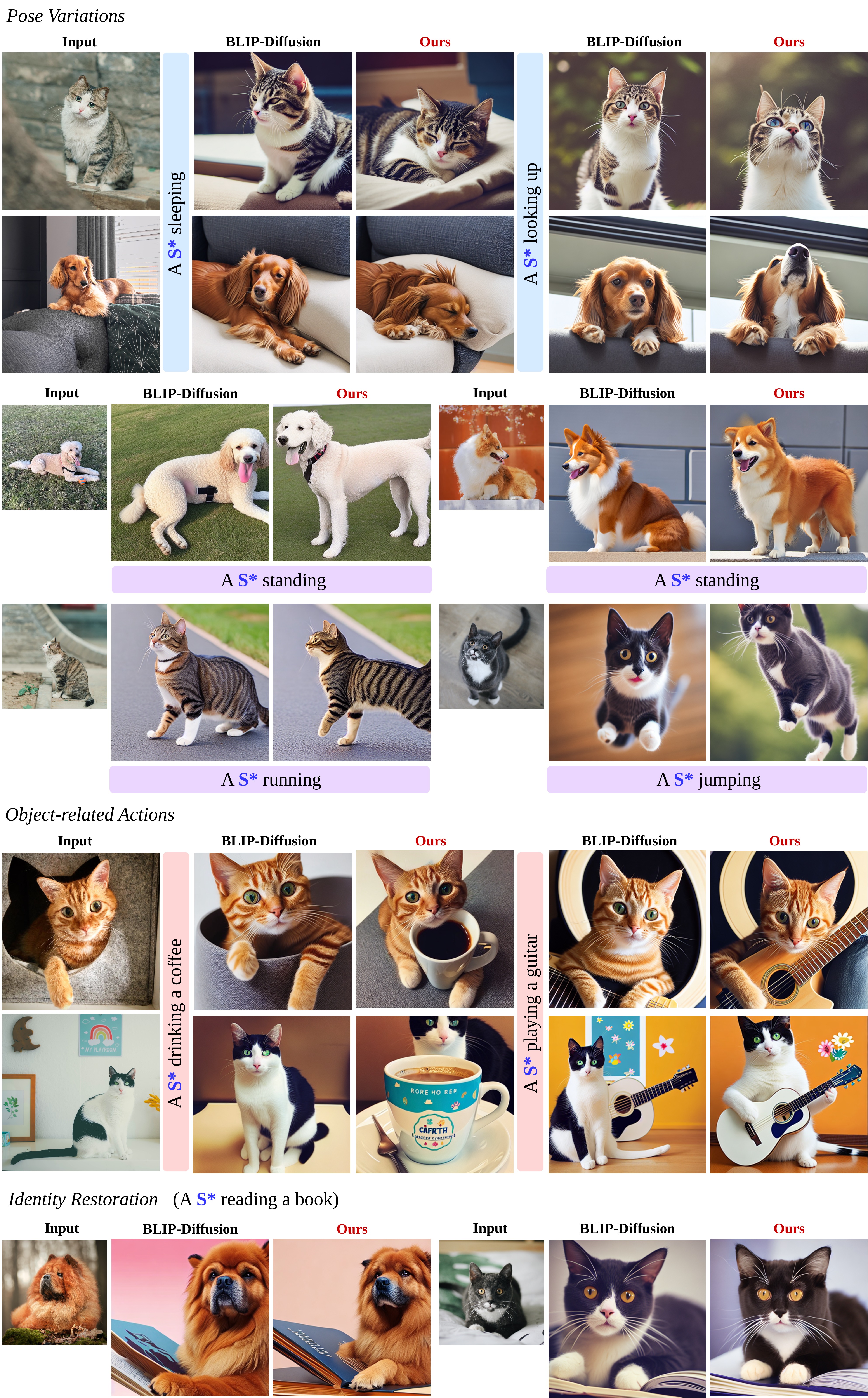}
    \caption{
    \textbf{Comparisons with BLIP-Diffusion.}
    }
    \label{fig:qualitative}
\end{figure}

\subsection{Qualitative Results}

We present a comparative analysis of our method with BLIP-Diffusion across three scenarios in \cref{fig:qualitative}.
In the 1st to 4th rows of \cref{fig:qualitative}, our method demonstrates the superior text alignment compared to the baseline when pose variation prompts are given. While the BLIP-Diffusion tends to replicate the subject's pose from the input image without adjustment, our method effectively modifies the pose in accordance with the provided prompt. Furthermore, we also found a similar tendency when using object-related prompts depicted in the 5th and 6th rows of \cref{fig:qualitative}.
It is noted that while the BLIP-Diffusion simply generates objects without executing the specified actions, our method demonstrates the capability to perform the appropriate action in response to the given prompt. In the perspective of subject's identity preservation, as shown in the last row of \cref{fig:qualitative}, ours preserve the identity better than the baseline.
We also demonstrate the improvements of our method compared to ELITE, particularly in terms of object-related action and subject identity preservation in \cref{fig:qualitative_elitr}. We provide more qualitative results in supplementary materials.

\begin{figure}[t!]
    \centering
    \includegraphics[width=\textwidth]{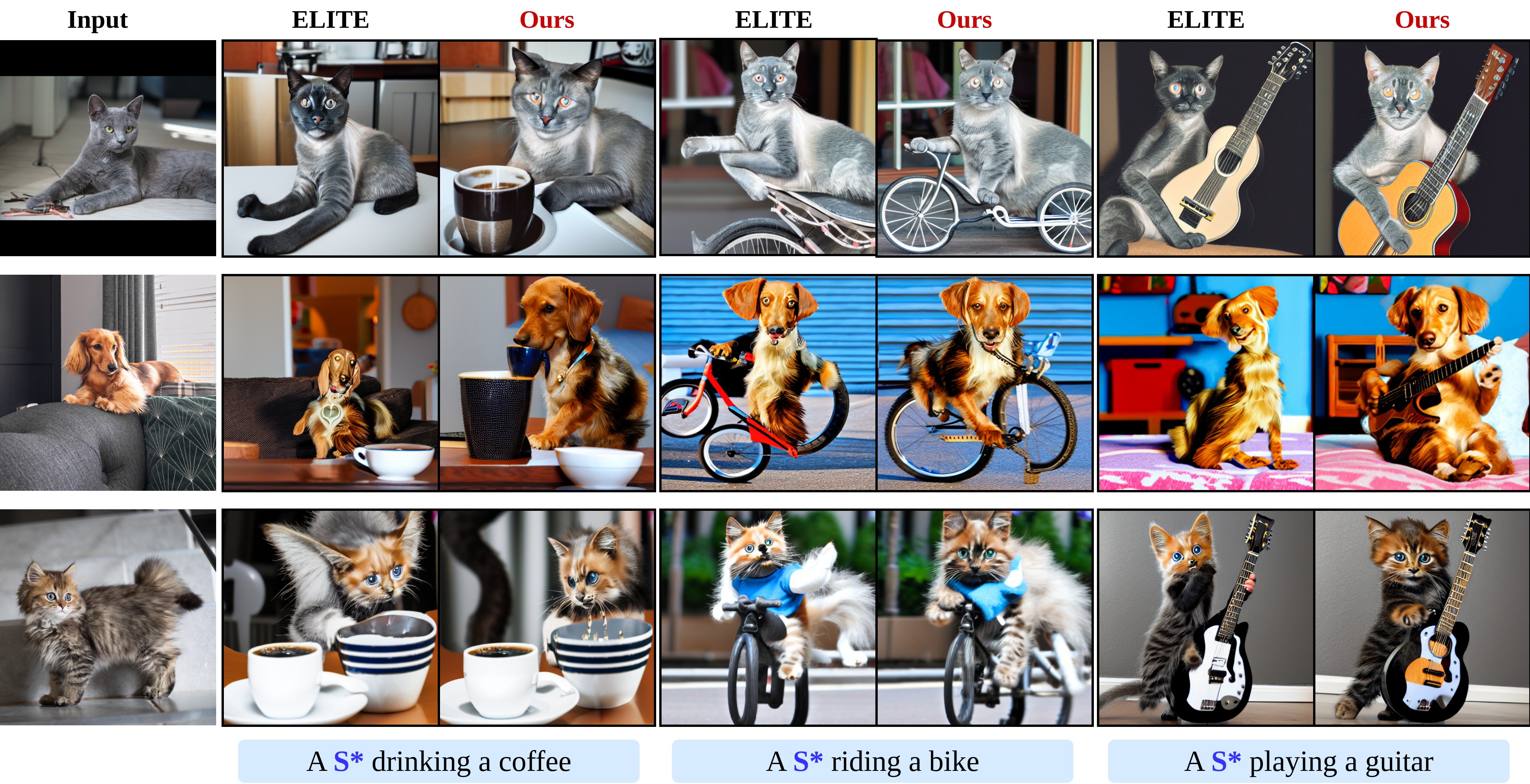}
    \caption{\textbf{Comparisons with ELITE.}}
    \label{fig:qualitative_elitr}
    \vspace{-1em}
\end{figure}

\subsection{Quantitative Results}

\cref{tab:benchmark} shows quantitative analyses using both the DS set and the DB set~\cite{ruiz2023dreambooth}.
In case of the DS set, which includes object-related action prompts as shown in 5--6th row of \cref{fig:qualitative}, we found that the newly generated object affects the image-alignment score, for example, by occluding the subject. Therefore, we additionally report metrics using the subject's segmentation masks (M-CLIP-I and M-DINO-I) to more properly evaluate the image alignment in a pose modification setting. For the DS set that consists of the prompts necessitating the deformation of the subject's pose, our method demonstrates significant improvement in the text alignment metric without compromising the image alignment metric compared to the baseline.
This suggests that our method effectively puts the subject in a pose according to the provided text prompt while preserving the subject's identity.
Furthermore, quantitative analysis on the DB set confirms that our methodology remains comparable to the existing baseline, even in typical scenarios where the subject is non-deformable object and the subject's pose is unchanged.

\begin{table}[t]
\caption{\textbf{Quantitative Comparison on Deformable Subject Set and DreamBooth dataset~\cite{ruiz2023dreambooth}.} `M-' indicates the metrics using segmentation masks.}
    \centering
	\resizebox{1\textwidth}{!}{
	\begin{tabular}{l|ccccc|ccc}
	\toprule
        & \multicolumn{5}{c|}{Deformable Subject Set} & \multicolumn{3}{c}{DreamBooth Set~\cite{ruiz2023dreambooth}} \\  
        \cmidrule(lr){2-9}
	{\centering Method} & {\centering \scriptsize \hskip0.5em CLIP-T($\uparrow$)} & {\centering \scriptsize \hskip1em M-CLIP-I($\uparrow$)} & {\centering \scriptsize \hskip1em M-DINO-I($\uparrow$)} & {\centering \scriptsize \hskip1em CLIP-I($\uparrow$)} & {\centering \scriptsize \hskip1em DINO-I($\uparrow$)} & {\centering \scriptsize \hskip0.5em CLIP-T($\uparrow$)} & {\centering \scriptsize \hskip1em CLIP-I($\uparrow$)} & {\centering \scriptsize \hskip1em DINO-I($\uparrow$)}\\
	\midrule
	BLIP-D~\cite{li2024blip} & 0.262 & 0.846 & 0.617 & \textbf{0.853} & 0.684 &  0.295 & 0.812 & 0.660 \\
	\textit{w/}Orchestration \hskip1em & \textbf{0.276} & 0.843 & 0.610 & 0.819 & 0.662 & \textbf{0.298} & 0.805 & 0.647\\
	\textit{w/}SA Swap & 0.259 & \textbf{0.848} & \textbf{0.631} & 0.834 & \textbf{0.693} & 0.293 & \textbf{0.815} & \textbf{0.675} \\
 	\midrule
	Ours & 0.272 & \textbf{0.848} & 0.630 & 0.819 & 0.681 & 0.296 & 0.809 & 0.665 \\
  	\midrule
        \midrule
	ELITE~\cite{wei2023elite} 
	& 0.285 & 0.835 & 0.574 &  0.751 & 0.486 & 0.294 &  0.792 &  0.664 \\
	\textit{w/}Orchestration \hskip1em & 0.292 & 0.834 & 0.570 & 0.754 & 0.493 & \textbf{0.295} &  0.789 &  0.658\\ 
	\textit{w/}SA Swap & 0.288 & \textbf{0.837} & \textbf{0.581} & 0.754 & 0.489 & 0.292 & \textbf{0.796} & \textbf{0.673}\\
  	\midrule
	Ours & \textbf{0.300} & \textbf{0.837} & \textbf{0.581} & \textbf{0.755} & \textbf{0.502} & 0.294 & 0.792 & 0.670 \\
	\bottomrule
	\end{tabular}}
    \label{tab:benchmark}
\vspace{-1em}
\end{table}

\subsection{User Study}

We further evaluate our method through the user study conducted with Amazon Mechanical Turk. Human raters were given a subject's input image, a prompt, and two synthesized images (ours and the baseline) of the subject. They were then asked to select the preferable one for each of the following questions: \textit{(1) Image Alignment}: ``Which of the images best reproduces the identity (\eg, item type and details) of the reference item?'' \textit{(2) Text Alignment}: ``Which of the images is best described by the reference text?''. We used 19 animal subjects and 11 pose-related prompts per subject, assigning 5 raters for each example.

\begin{wrapfigure}{r}{5cm}
    \centering
    \includegraphics[width=0.4\textwidth]{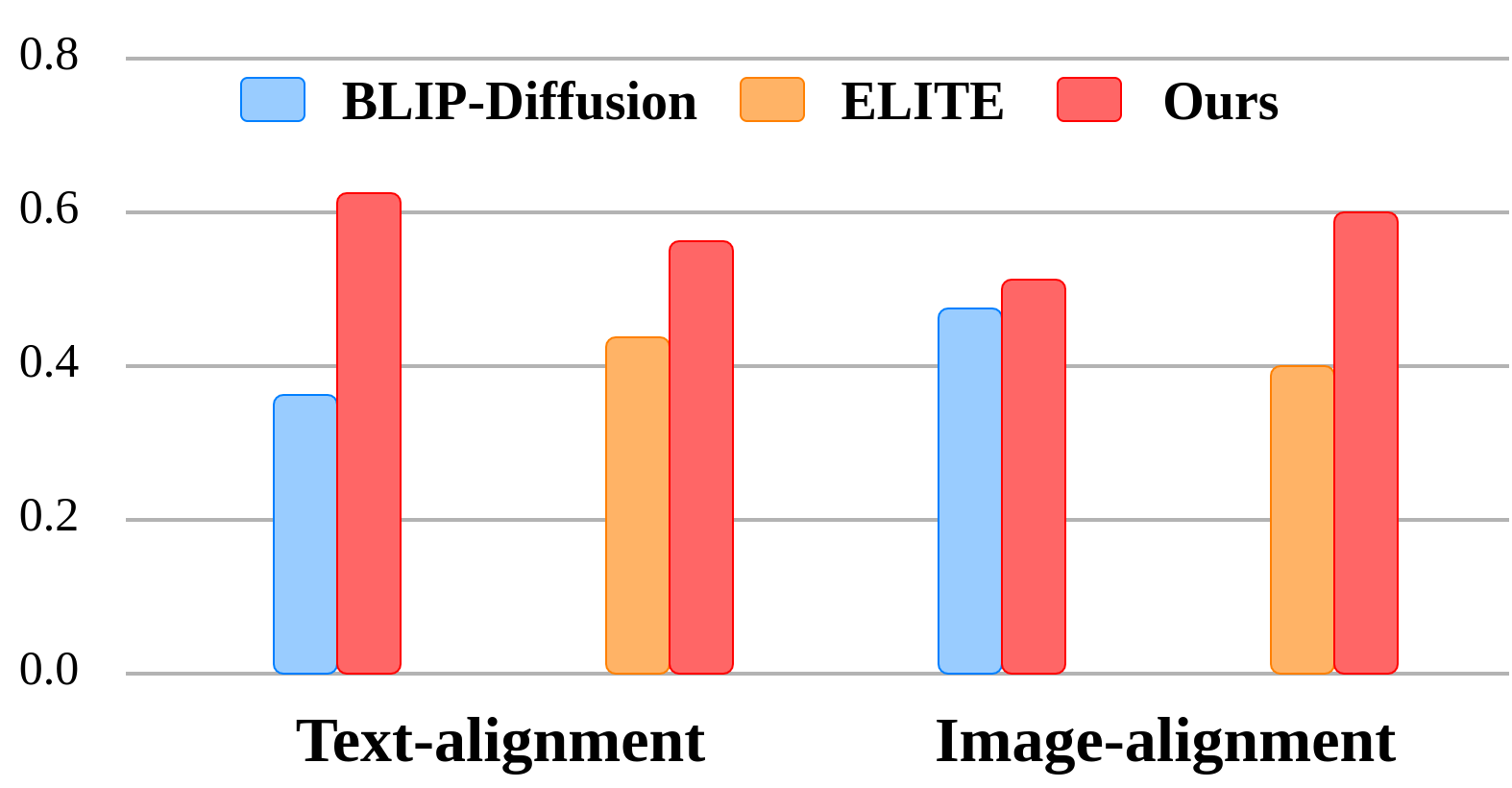}
    \captionof{figure}{\textbf{User Study Results}.}
    \label{fig:userstudy}
\end{wrapfigure}

As shown in \cref{fig:userstudy}, our method is preferred over the baseline in both text alignment (Ours 62.0\% vs. BLIP-Diffusion 38.0\%, Ours 56.3\% vs. ELITE 43.7\%) and image alignment (Ours 51.5\% vs. BLIP-Diffusion 48.5\%, Ours 60.0\% vs. ELITE 40.0\%). It is noteworthy that our method was selected for the better alignment of both text and image,
underscoring its ability to faithfully follow the text prompt and preserve the subject's identity from a human perspective.

\vspace{-0.5em}
\subsection{Ablations}
\begin{figure}[t]
    \centering
    \includegraphics[width=\textwidth]{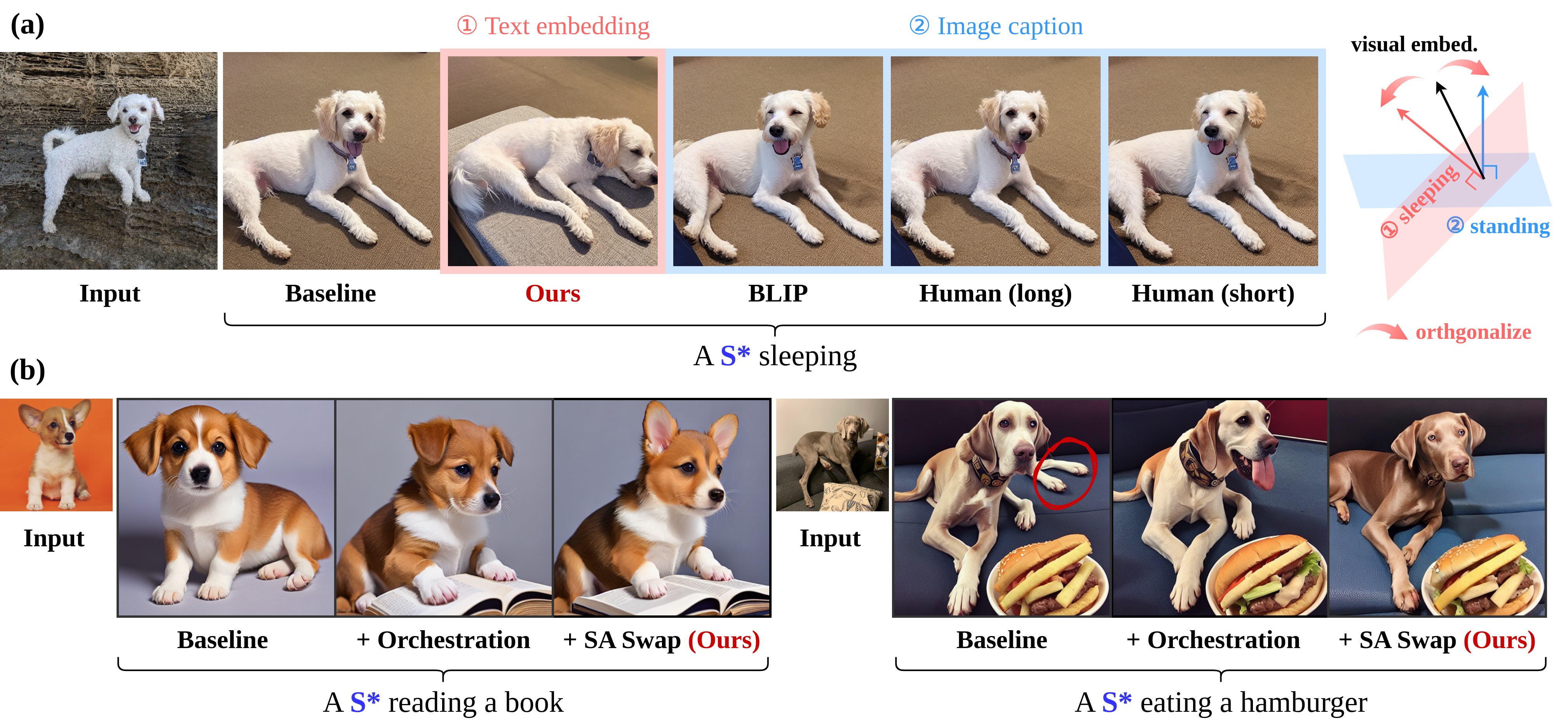}
    \caption{\textbf{Ablation Results}. (a) We compare different orthogonalizations with respect to various textual embeddings,  derived from the input image's caption and text prompt (ours). Explicitly considering a relationship between the input image and the text prompt effectively aligns the visual embedding.
    (b) We verify the effectiveness of each component in our method. Orchestration encourages generated images to faithfully follow the text prompt while self-attention swap successfully retains the subject's identity.} 
    \label{fig:ablation}
    \vspace{-1.5em}
\end{figure}

Our orchestration method eliminates the conflicting elements in the visual embedding \textit{that interferes with the textual embedding.}
Then, a subsequent question arises: why don't we directly orthogonalize the visual embedding with respect to the text description about the input image? 
In other words, why not eliminate such elements by only accessing the subject’s image without considering the accompanying text prompt embedding? To investigate the effect of orthogonalization with respect to 
the embedding vectors obtained from (i) the text prompt for generation and (ii) the caption describing the input image (\cref{fig:ablation}(a)), we conduct the following experiment: we adopt the existing image captioning model~\cite{li2022blip} and transform the visual embedding to be orthogonal to its caption's embedding using \cref{eq:orthogonal}. 
We also use human annotations that describe the input image in both short and long versions. 
\cref{fig:ablation}(a) shows that without considering the text prompt embedding, the generated images are strongly biased toward the input image. 
More concretely, when we orthogonalize the visual embedding with respect to the embedding vectors derived from the captions describing the input image, like BLIP captioning 
(``A dog standing on a rock'') 
or 
human annotations,
(``A dog  standing'' and ``A dog standing upright on a rock, facing forward, with its mouth slightly open and holding the right front paw a bit elevated'')
the discord among the contextual embeddings remains unalleviated. On the other hand, our method of orthogonalization with respect to the text embedding effectively resolves the discord, rendering the images faithful to the text prompt without requiring any additional language model or human endeavor.
% With our contextual embedding orchestration, we are able to generate images faithful to the text prompt effectively and efficiently.

We also conducted an ablation study on the roles of each component, orchestration and self-attention swap, by progressively applying them. As shown in \cref{fig:ablation}(b) and \cref{tab:benchmark}, orchestration alone makes huge progress on faithfully following the text prompt, while loss in the subject's identity remains unsolved. After applying self-attention swap which restores the subject's identity, our proposed method successfully generates images that accurately adhere to the text prompt while maintaining the subject's identity.

\vspace{-0.5em}
\subsection{Analysis}

We conduct the same experiment as in \cref{subsec:discord} with our method and report the result. Red dots in \cref{fig:discord}(b) shows that with the orchestration of the contextual embedding, the image features of the generated images are properly aligned between two image features \ie, the outputs when solely using visual embedding $\text{c}_{\textbf{v},\varnothing}$ and solely using textual embedding $\text{c}_{\varnothing,\textbf{c}}$, respectively. We also computed LPIPS~\cite{zhang2018unreasonable} distance of our generated images with the outputs using the visual-only embedding (LPIPS=6.78) and with the outputs using the textual-only embedding (LPIPS=6.64), respectively. Compared to the baseline (LPIPS=5.69, 7.25) remarkably closer to the visual-only outputs, the contextual embedding orchestration put the generated images in appropriate points between identity preservation and text alignment. Note that an increase in LPIPS distance with the outputs from the visual-only embedding is an expected consequence of a change in the subject's context (\eg, pose).

\begin{figure}[t]
    \centering
    \includegraphics[width=\textwidth]{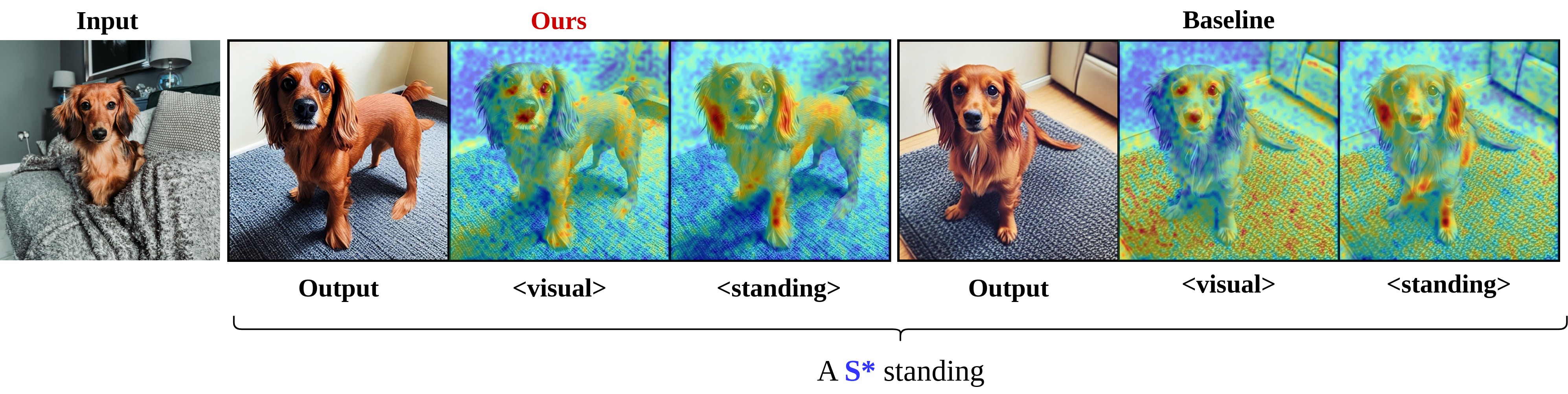}
    \caption{\textbf{Visualization of each token's cross-attention map} (red: high value, blue: low value). In baseline, the visual and textual embeddings interfere with each other, attending to irrelevant areas (\eg, beneath the dog). Meanwhile, our method effectively resolves the discord among the contextual embedding, injecting each visual and text information in the proper areas. }
    \vspace{-1.5em}
    \label{fig:analysis}
\end{figure}

To verify that the visual embedding is properly adjusted after orchestration, we also visualized cross-attention maps for tokens corresponding to the visual and textual embeddings. 
Cross-attention maps indicate the amount of information conveyed from the tokens to the latent spatial features $f$, therefore, we can obtain insight about the information flows of each token. 
We interpolated cross-attention maps from each layer in both the encoder and decoder, averaging them across layers and tokens. 
As shown in \cref{fig:analysis}, in the baseline, the discord among the contextual embedding leads the textual embedding (``<standing>'') to highlight the irrelevant areas (\eg, beneath the dog), resulting in the image noncompliant with the text prompt. Meanwhile, our orchestration effectively resolves the discord, integrating the textual embedding in the proper areas. 

\section{Conclusion}

\textbf{Limitations.} Our method inherits the weakness from T2I diffusion models and falls short of complying several text prompts at once. When a set of prompts in different instructions is given, our method cannot accurately orthogonalize the visual embedding to each textual embedding, resulting in the images omitting some instructions. We provide more details in supplementary materials.

In this paper, we explore notable problems in pose variation tasks when using the zero-shot customization methods. We focus on a conflict between the visual and textual embeddings, which affects both embeddings and leads to the \textit{pose bias} and \textit{identity loss}, respectively. We alleviate the conflict by \textit{contextual embedding orchestratation}, that is, orthogonalize the visual embedding with respect to the textual embedding subspace. While orchestration effectively put the subject in various poses complying with the text prompts, the subject's identity remains partially damaged. To resolve the issue, we propose \textit{self-attention swap}, that adopts the subject's clean identity from the visual-only embedding to the appropriate latent pixels. Amidst the proliferation of zero-shot customization methods and the utilization of the visual embedding, our key insight unveils a crucial aspect for achieving a more diverse and pose-variant generation.

%p.14
\clearpage  % TODO REVIEW/FINAL: This \clearpage needs to be removed from both review and camera-ready versions.

\bibliographystyle{splncs04}
\bibliography{main}

\begin{thebibliography}{10}
\providecommand{\url}[1]{\texttt{#1}}
\providecommand{\urlprefix}{URL }
\providecommand{\doi}[1]{https://doi.org/#1}

\bibitem{armandpour2023perpneg}
Armandpour, M., Sadeghian, A., Zheng, H., Sadeghian, A., Zhou, M.: Re-imagine the negative prompt algorithm: Transform 2d diffusion into 3d, alleviate janus problem and beyond (2023)

\bibitem{avrahami2023break}
Avrahami, O., Aberman, K., Fried, O., Cohen-Or, D., Lischinski, D.: Break-a-scene: Extracting multiple concepts from a single image. arXiv preprint arXiv:2305.16311  (2023)

\bibitem{Avrahami_2022blendeddiffusion}
Avrahami, O., Lischinski, D., Fried, O.: Blended diffusion for text-driven editing of natural images. In: 2022 IEEE/CVF Conference on Computer Vision and Pattern Recognition (CVPR). IEEE (Jun 2022). \doi{10.1109/cvpr52688.2022.01767}, \url{http://dx.doi.org/10.1109/CVPR52688.2022.01767}

\bibitem{balaji2023ediffi}
Balaji, Y., Nah, S., Huang, X., Vahdat, A., Song, J., Zhang, Q., Kreis, K., Aittala, M., Aila, T., Laine, S., Catanzaro, B., Karras, T., Liu, M.Y.: ediff-i: Text-to-image diffusion models with an ensemble of expert denoisers (2023)

\bibitem{cao_2023_masactrl}
Cao, M., Wang, X., Qi, Z., Shan, Y., Qie, X., Zheng, Y.: Masactrl: Tuning-free mutual self-attention control for consistent image synthesis and editing. In: Proceedings of the IEEE/CVF International Conference on Computer Vision (ICCV). pp. 22560--22570 (October 2023)

\bibitem{chen2023photoverse}
Chen, L., Zhao, M., Liu, Y., Ding, M., Song, Y., Wang, S., Wang, X., Yang, H., Liu, J., Du, K., Zheng, M.: Photoverse: Tuning-free image customization with text-to-image diffusion models (2023)

\bibitem{couairon2022diffedit}
Couairon, G., Verbeek, J., Schwenk, H., Cord, M.: Diffedit: Diffusion-based semantic image editing with mask guidance (2022)

\bibitem{gal2022textual_inversion}
Gal, R., Alaluf, Y., Atzmon, Y., Patashnik, O., Bermano, A.H., Chechik, G., Cohen-Or, D.: An image is worth one word: Personalizing text-to-image generation using textual inversion (2022)

\bibitem{gal2022image}
Gal, R., Alaluf, Y., Atzmon, Y., Patashnik, O., Bermano, A.H., Chechik, G., Cohen-Or, D.: An image is worth one word: Personalizing text-to-image generation using textual inversion. arXiv preprint arXiv:2208.01618  (2022)

\bibitem{hertz2022prompt}
Hertz, A., Mokady, R., Tenenbaum, J., Aberman, K., Pritch, Y., Cohen-Or, D.: Prompt-to-prompt image editing with cross attention control. arXiv preprint arXiv:2208.01626  (2022)

\bibitem{ho2020ddpm}
Ho, J., Jain, A., Abbeel, P.: Denoising diffusion probabilistic models (2020)

\bibitem{jia2023tamingencoder}
Jia, X., Zhao, Y., Chan, K.C.K., Li, Y., Zhang, H., Gong, B., Hou, T., Wang, H., Su, Y.C.: Taming encoder for zero fine-tuning image customization with text-to-image diffusion models (2023)

\bibitem{karras2022elucidating}
Karras, T., Aittala, M., Aila, T., Laine, S.: Elucidating the design space of diffusion-based generative models. Advances in Neural Information Processing Systems  \textbf{35},  26565--26577 (2022)

\bibitem{kawar2023imagic}
Kawar, B., Zada, S., Lang, O., Tov, O., Chang, H., Dekel, T., Mosseri, I., Irani, M.: Imagic: Text-based real image editing with diffusion models (2023)

\bibitem{kim2023dense}
Kim, Y., Lee, J., Kim, J.H., Ha, J.W., Zhu, J.Y.: Dense text-to-image generation with attention modulation (2023)

\bibitem{kumari2023multi}
Kumari, N., Zhang, B., Zhang, R., Shechtman, E., Zhu, J.Y.: Multi-concept customization of text-to-image diffusion. In: Proceedings of the IEEE/CVF Conference on Computer Vision and Pattern Recognition. pp. 1931--1941 (2023)

\bibitem{li2024blip}
Li, D., Li, J., Hoi, S.: Blip-diffusion: Pre-trained subject representation for controllable text-to-image generation and editing. Advances in Neural Information Processing Systems  \textbf{36} (2024)

\bibitem{li2022blip}
Li, J., Li, D., Xiong, C., Hoi, S.: Blip: Bootstrapping language-image pre-training for unified vision-language understanding and generation. In: International Conference on Machine Learning. pp. 12888--12900. PMLR (2022)

\bibitem{liu2022pseudo}
Liu, L., Ren, Y., Lin, Z., Zhao, Z.: Pseudo numerical methods for diffusion models on manifolds. arXiv preprint arXiv:2202.09778  (2022)

\bibitem{liu2023compositional}
Liu, N., Li, S., Du, Y., Torralba, A., Tenenbaum, J.B.: Compositional visual generation with composable diffusion models (2023)

\bibitem{ma2023subjectdiffusionopen}
Ma, J., Liang, J., Chen, C., Lu, H.: Subject-diffusion:open domain personalized text-to-image generation without test-time fine-tuning (2023)

\bibitem{mokady2022nulltext}
Mokady, R., Hertz, A., Aberman, K., Pritch, Y., Cohen-Or, D.: Null-text inversion for editing real images using guided diffusion models (2022)

\bibitem{mou2023t2iadapter}
Mou, C., Wang, X., Xie, L., Wu, Y., Zhang, J., Qi, Z., Shan, Y., Qie, X.: T2i-adapter: Learning adapters to dig out more controllable ability for text-to-image diffusion models (2023)

\bibitem{nichol2022glide}
Nichol, A., Dhariwal, P., Ramesh, A., Shyam, P., Mishkin, P., McGrew, B., Sutskever, I., Chen, M.: Glide: Towards photorealistic image generation and editing with text-guided diffusion models (2022)

\bibitem{parmar2023pix2pix_zero}
Parmar, G., Singh, K.K., Zhang, R., Li, Y., Lu, J., Zhu, J.Y.: Zero-shot image-to-image translation (2023)

\bibitem{qiu2023controlling}
Qiu, Z., Liu, W., Feng, H., Xue, Y., Feng, Y., Liu, Z., Zhang, D., Weller, A., Schölkopf, B.: Controlling text-to-image diffusion by orthogonal finetuning (2023)

\bibitem{radford2021clip}
Radford, A., Kim, J.W., Hallacy, C., Ramesh, A., Goh, G., Agarwal, S., Sastry, G., Askell, A., Mishkin, P., Clark, J., Krueger, G., Sutskever, I.: Learning transferable visual models from natural language supervision (2021)

\bibitem{ramesh2022dalle2}
Ramesh, A., Dhariwal, P., Nichol, A., Chu, C., Chen, M.: Hierarchical text-conditional image generation with clip latents (2022)

\bibitem{rombach2022ldm}
Rombach, R., Blattmann, A., Lorenz, D., Esser, P., Ommer, B.: High-resolution image synthesis with latent diffusion models (2022)

\bibitem{ruiz2023dreambooth}
Ruiz, N., Li, Y., Jampani, V., Pritch, Y., Rubinstein, M., Aberman, K.: Dreambooth: Fine tuning text-to-image diffusion models for subject-driven generation. In: Proceedings of the IEEE/CVF Conference on Computer Vision and Pattern Recognition. pp. 22500--22510 (2023)

\bibitem{saharia2022imagen}
Saharia, C., Chan, W., Saxena, S., Li, L., Whang, J., Denton, E., Ghasemipour, S.K.S., Ayan, B.K., Mahdavi, S.S., Lopes, R.G., Salimans, T., Ho, J., Fleet, D.J., Norouzi, M.: Photorealistic text-to-image diffusion models with deep language understanding (2022)

\bibitem{shi2023instantbooth}
Shi, J., Xiong, W., Lin, Z., Jung, H.J.: Instantbooth: Personalized text-to-image generation without test-time finetuning (2023)

\bibitem{simonyan2014very}
Simonyan, K., Zisserman, A.: Very deep convolutional networks for large-scale image recognition. arXiv preprint arXiv:1409.1556  (2014)

\bibitem{sohldickstein2015deep}
Sohl-Dickstein, J., Weiss, E.A., Maheswaranathan, N., Ganguli, S.: Deep unsupervised learning using nonequilibrium thermodynamics (2015)

\bibitem{tewel2023perfusion}
Tewel, Y., Gal, R., Chechik, G., Atzmon, Y.: Key-locked rank one editing for text-to-image personalization (2023)

\bibitem{tewel2023key}
Tewel, Y., Gal, R., Chechik, G., Atzmon, Y.: Key-locked rank one editing for text-to-image personalization. In: ACM SIGGRAPH 2023 Conference Proceedings. pp. 1--11 (2023)

\bibitem{tumanyan2022plugandplay}
Tumanyan, N., Geyer, M., Bagon, S., Dekel, T.: Plug-and-play diffusion features for text-driven image-to-image translation (2022)

\bibitem{voynov2023p}
Voynov, A., Chu, Q., Cohen-Or, D., Aberman, K.: P+: Extended textual conditioning in text-to-image generation (2023)

\bibitem{wei2023elite}
Wei, Y., Zhang, Y., Ji, Z., Bai, J., Zhang, L., Zuo, W.: Elite: Encoding visual concepts into textual embeddings for customized text-to-image generation. arXiv preprint arXiv:2302.13848  (2023)

\bibitem{xiao2023fastcomposer}
Xiao, G., Yin, T., Freeman, W.T., Durand, F., Han, S.: Fastcomposer: Tuning-free multi-subject image generation with localized attention (2023)

\bibitem{yang2022paint}
Yang, B., Gu, S., Zhang, B., Zhang, T., Chen, X., Sun, X., Chen, D., Wen, F.: Paint by example: Exemplar-based image editing with diffusion models (2022)

\bibitem{yuan2023customnet}
Yuan, Z., Cao, M., Wang, X., Qi, Z., Yuan, C., Shan, Y.: Customnet: Zero-shot object customization with variable-viewpoints in text-to-image diffusion models (2023)

\bibitem{zhang2023controlnet}
Zhang, L., Rao, A., Agrawala, M.: Adding conditional control to text-to-image diffusion models (2023)

\bibitem{zhang2018unreasonable}
Zhang, R., Isola, P., Efros, A.A., Shechtman, E., Wang, O.: The unreasonable effectiveness of deep features as a perceptual metric. In: Proceedings of the IEEE conference on computer vision and pattern recognition. pp. 586--595 (2018)

\end{thebibliography}
\end{document}

\title{Supplementary Materials for \\ Harmonizing Visual and Textual Embeddings \\ for Zero-Shot Text-to-Image Customization} 

\titlerunning{Abbreviated paper title}

\author{First Author\inst{1}\orcidlink{0000-1111-2222-3333} \and
Second Author\inst{2,3}\orcidlink{1111-2222-3333-4444} \and
Third Author\inst{3}\orcidlink{2222--3333-4444-5555}}

\authorrunning{F.~Author et al.}

\institute{Princeton University, Princeton NJ 08544, USA \and
Springer Heidelberg, Tiergartenstr.~17, 69121 Heidelberg, Germany
\email{lncs@springer.com}\\
\url{http://www.springer.com/gp/computer-science/lncs} \and
ABC Institute, Rupert-Karls-University Heidelberg, Heidelberg, Germany\\
\email{\{abc,lncs\}@uni-heidelberg.de}}

\pagenumbering{roman}
\renewcommand\thesection{\Alph{section}}
\renewcommand\thetable{\roman{table}}
\renewcommand\thefigure{\roman{figure}}
\setcounter{table}{0}
\setcounter{figure}{0}

\maketitle

\section{Problems in Pose Modification}

We identify two problems when modifying the subject's pose with a text prompt: \textit{pose bias} and \textit{identity loss}. %Due to the pose-identity entanglement, the visual embedding includes the subject's pose information in the input image, resulting in a conflict with the textual embedding that carries novel pose information. 
As shown in \cref{fig:suppl_problems}, the results from visual-only embedding and textual-only embedding exhibit high subject fidelity and text alignment, respectively, while using both embedding leads to a conflict and generates images showing the two problems. 

\begin{figure}[h!]
\centering
  \includegraphics[width=1\linewidth]{fig/Supplement_problems.png}\\
  \caption{\textbf{Problems in Pose Modification.}}
  \label{fig:suppl_problems}
\end{figure}

\cref{fig:suppl_identity} illustrates the cases when the identity loss appears in the baseline~\cite{li2024blip}. When the textual embedding interferes with the visual embedding, it also affects the subject's identity, damaging the subject's essential features \eg, face color or fur pattern. Meanwhile, our method effectively restores the subject's identity while faithfully following the text prompt by adopting self-attention swap that leverages the clear identity information from the visual-only embedding. We also conduct the experiment generating images with visual-only embedding and species-textual-only embedding, respectively. For species-textual-only embedding, we compose the contextual embedding only with the text embedding consisting of the species name of the subject and the subject's pose in the input image. Therefore, the species-textual-only embedding results in images that share the same species and pose with the subject in the input image, while not including the subject's unique identity. We compare LPIPS~\cite{zhang2018unreasonable} distance of the images from the baseline with the outputs using the visual-only embedding (LPIPS=6.44) and with the outputs using the species-textual-only embedding (LPIPS=2.44). Compared to our method (LPIPS=2.18, 2.44), the baseline generates the subject resembling the general species rather than containing the subject's identity, indicating the identity loss.

\begin{figure}[h!]
\centering
  \includegraphics[width=1\linewidth]{fig/Supplement_identity3.png}\\
  \caption{\textbf{Comparisons Regarding Identity Loss.}}
  \label{fig:suppl_identity}
\end{figure}

\section{Implementation Details}

\textbf{Self-attention Swap.} Following the analysis in \cite{cao_2023_masactrl}, we applied self-attention swap in the layers 10--15 in the decoder. We obtain a binary mask $m$ from the cross-attention maps $M_C$ that associate with the class word for \cite{li2024blip} and pseudo-token for\cite{wei2023elite}, using a threshold $\tau=0.5$.

\section{Experimental Details}

\textbf{Masked Scores.} In our experiments, Deformable Subjet Set (DS set) comprises prompts that change the subject's pose while sometimes actively interacting with the specific object (\eg, A $\text{S}^*$ playing a guitar). However, measuring the subject's fidelity with CLIP-I and DINO-I, as proposed in DreamBooth~\cite{ruiz2023dreambooth}, leads to insufficient evaluation for these outputs since the newly generated object affects the score. Therefore, we additionally adopt the masked scores for CLIP-I and DINO-I using the subject's segmentation mask for both the input images (or reference) and generated images (or output). We used Grounded-SAM~\cite{ren2024grounded} for obtaining the segmentation mask using the subject's class as the prompt. \cref{fig:suppl_metrics} indicates that when a prompt is associated with the newly introduced object, masked scores offer a more accurate assessment, mitigating the impact of the new object.

\begin{figure}[h!]
\centering
  \includegraphics[width=1\linewidth]{fig/Supplement_masked_metrics2.png}\\
  \caption{\textbf{Comparisons between image alignment scores.} We evaluate the image alignment of the generated images from a text prompt, \textbf{A \textcolor{blue}{$\text{S}^*$} is playing a guitar}. In the 1st column, the image in the 2nd row exhibits better identity preservation, while obtaining the lower score due to the newly generated object, a guitar. As shown in the 4th column, adopting the segmentation masks for both the reference and the outputs leads to a more accurate assessment, mitigating the impact of the object.}
  \label{fig:suppl_metrics}
\end{figure}

\bigskip
\noindent \textbf{Deformable Subject Set.} To evaluate the model's ability to faithfully modify the subject's pose while preserving its identity, we compose Deformable Subject Set (DS set) by collecting 19 animals from DreamBench~\cite{ruiz2023dreambooth} and CustomDiffusion~\cite{kumari2023multi}. \cref{fig:suppl_ds_set} represents the full input images of the subjects. For quantitative evaluation, we employ 11 prompts that instruct pose modifications for each subject, as listed in \cref{tab:suppl_ds_set}.

\begin{figure}[h!]
\centering
  \includegraphics[width=1\linewidth]{fig/Supplement_ds_set.jpg}\\
  \caption{\textbf{The DS set image samples for quantitative evaluation.}}
  \label{fig:suppl_ds_set}
\end{figure}

\begin{table}[h!]
\caption{\textbf{Text Prompts for quantitative evaluation in the DS set.}}
\setlength{\tabcolsep}{10pt}
\renewcommand{\arraystretch}{1.5}
    \centering
    {\footnotesize
    \begin{tabular}{c|c|c}
    \hline
        a $\text{S}^*$ drinking a coffee & a $\text{S}^*$ eating a hamburger & a $\text{S}^*$ flying \\ 
        a $\text{S}^*$ jumping & a $\text{S}^*$ lying on the sofa & a $\text{S}^*$ playing a guitar \\ 
        a $\text{S}^*$ reading a book & a $\text{S}^*$ riding a bike & a $\text{S}^*$ running \\ 
        a $\text{S}^*$ sleeping & a $\text{S}^*$ standing & \\
    \hline
    \end{tabular}
    }
    \label{tab:suppl_ds_set}
\end{table}

\section{Limitation}

Inheriting a characteristic from T2I diffusion models, our method faces difficulties in complying with multiple prompts simultaneously. As shown in the 1st row of \cref{fig:suppl_limitation}, both BLIP-Diffusion~\cite{li2024blip} and our method fall short of applying multiple prompts aimed at generating different surroundings. Meanwhile, our method exhibits relatively more robustness in following the prompts that change the subject's pose, as shown in the 2nd row of \cref{fig:suppl_limitation}. However, this limitation in both methods would provide a glimpse into avenues for improving zero-shot customization in future works.

\begin{figure}[h!]
\centering
  \includegraphics[width=1\linewidth]{fig/Supplement_limitation.jpg}\\
  \caption{\textbf{Limitation of our method.} Both our method and the baseline~\cite{li2024blip} fail to follow the multiple prompts simultaneously. Failed prompts are in strikethrough above.}
  \label{fig:suppl_limitation}
\end{figure}

\section{More Qualitative Results}

We provide more qualitative results in \cref{fig:suppl_qualitative_1,fig:suppl_qualitative_2,fig:suppl_qualitative_3,fig:suppl_qualitative_4,fig:suppl_qualitative_5,fig:suppl_qualitative_6,fig:suppl_qualitative_7,fig:suppl_qualitative_8}. We show the input image in the 1st column and provide text prompts under each image.

\clearpage

\begin{figure}[t]
\centering
  \includegraphics[width=1\linewidth]{fig/Supplement_qualitative_cat__0.jpg}\\
  \caption{\textbf{Additional Comparison Results with the Baseline~\cite{li2024blip}.}}
  \label{fig:suppl_qualitative_1}
\end{figure}

\begin{figure}[t]
\centering
  \includegraphics[width=1\linewidth]{fig/Supplement_qualitative_dog6__4.jpg}\\
  \caption{\textbf{Additional Comparison Results with the Baseline~\cite{li2024blip}.}}
  \label{fig:suppl_qualitative_2}
\end{figure}

\begin{figure}[t]
\centering
  \includegraphics[width=1\linewidth]{fig/Supplement_qualitative_pet_cat_5__8.jpg}\\
  \caption{\textbf{Additional Comparison Results with the Baseline~\cite{li2024blip}.}}
    \label{fig:suppl_qualitative_3}
\end{figure}

\begin{figure}[t]
\centering
  \includegraphics[width=1\linewidth]{fig/Supplement_qualitative_dog3__5.jpg}\\
  \caption{\textbf{Additional Comparison Results with the Baseline~\cite{li2024blip}.}}
  \label{fig:suppl_qualitative_4}
\end{figure}

\begin{figure}[t]
\centering
  \includegraphics[width=1\linewidth]{fig/Supplement_qualitative_dog_2__0.jpg}\\
  \caption{\textbf{Additional Comparison Results with the Baseline~\cite{li2024blip}.}}
  \label{fig:suppl_qualitative_5}
\end{figure}

\begin{figure}[t]
\centering
  \includegraphics[width=1\linewidth]{fig/Supplement_qualitative_elite.jpg}\\
  \caption{\textbf{Additional Comparison Results with the Baseline~\cite{wei2023elite}.}}
  \label{fig:suppl_qualitative_6}
\end{figure}

\begin{figure}[t]
\centering
  \includegraphics[width=1\linewidth]{fig/Supplement_qualitative_bias.jpg}\\
  \caption{\textbf{Qualitative Results using obfuscated input images.} When the input image includes the subject in a distinct pose that makes it difficult to discern the full body structures, the contextual embeddings in the baseline severely interfere with each other. As a result, the subject is depicted with the predominant body parts while aligning awkwardly with the text prompt (in the 1st -- 2nd row) or partially loses its identity (in the 3rd row). In contrast, our method alleviates the interference and swaps the subject's clean identity, satisfying both the text prompt and the subject fidelity.}
  \label{fig:suppl_qualitative_7}
\end{figure}

\begin{figure}[t]
\centering
  \includegraphics[width=1\linewidth]{fig/Supplement_qualitative_random.jpg}\\
  \caption{\textbf{Qualitative Results Generated from 3 random seeds.} To show our method's robustness, we randomly select 3 outputs from different seeds. Our method effectively and consistently changes the subject's pose across various subjects and prompts.}
  \label{fig:suppl_qualitative_8}
\end{figure}

\clearpage  % TODO REVIEW/FINAL: This \clearpage needs to be removed from both review and camera-ready versions.

\bibliographystyle{splncs04}
\bibliography{main}